\documentclass{llncs}
\usepackage{makeidx}  
\usepackage{csquotes}
\usepackage{amsmath}
\usepackage{algorithm}
\usepackage[pdftex]{graphicx} 
\usepackage[noend]{algpseudocode}
\begin{document}
\frontmatter          
\pagestyle{headings}  

\mainmatter              
\title{CFGs-2-NLU: \\ Sequence-to-Sequence Learning for Mapping Utterances to Semantics and Pragmatics}
\titlerunning{seq2seq for text to dialogue}  
\author{Adam James Summerville \and James Ryan \and
Michael Mateas \and \\ Noah Wardrip-Fruin}
\authorrunning{Adam James Summerville et al.} 
\institute{University of California, Santa Cruz \\
 1156 High Street \\
Santa Cruz, CA 95066 \\
\email{asummerv@ucsc.edu}, \email{\{jor,michaelm,nwf\}@soe.ucsc.edu}}

\maketitle              

\begin{abstract}

In this paper, we present a novel approach to natural language understanding that utilizes \textit{context-free grammars} (CFGs) in conjunction with \textit{sequence-to-sequence} (seq2seq) \textit{deep learning}. Specifically, we take a CFG authored to generate dialogue for our target application for NLU, a videogame, and train a \textit{long short-term memory} (LSTM) \textit{recurrent neural network} (RNN) to map the surface utterances that it produces to traces of the grammatical expansions that yielded them. Critically, this CFG was authored using a tool we have developed that supports arbitrary annotation of the nonterminal symbols in the grammar. Because we already annotated the symbols in this grammar for the semantic and pragmatic considerations that our game's dialogue manager operates over, we can use the grammatical trace associated with any surface utterance to infer such information. During gameplay, we translate player utterances into grammatical traces (using our RNN), collect the mark-up attributed to the symbols included in that trace, and pass this information to the dialogue manager, which updates the conversation state accordingly. From an offline evaluation task, we demonstrate that our trained RNN translates surface utterances to grammatical traces with great accuracy. To our knowledge, this is the first usage of seq2seq learning for conversational agents (our game's characters) who explicitly reason over semantic and pragmatic considerations.
  
\end{abstract}
\section{Introduction}

Conversational agents have become an increasingly common part of everyday life, with many service sector interactions being handled either by spoken dialogue or chat-based systems. But while service dialogue systems have become common, general conversational agents are still an open area of research. This is especially the case in entertainment-based interactive media, such as videogames (our targeted application domain, as we explain below), where only \textit{Fa\c{c}ade} has featured freeform conversational interactions with non-player characters (NPCs) \cite{mateas2004natural}. Due to the highly structured nature of the interactions in  service dialogue systems, rule-based systems for natural language understanding (NLU) can be successful in that interaction paradigm, e.g., by pattern matching of the form:
\begin{displayquote}
\textbf{Human}: I need to travel from San Francisco to New York on June 27th
\textbf{Computer}: * travel from \textit{[LOCATION]} to \textit{[LOCATION]} on \textit{[DATE]}
\end{displayquote}
But in conversational applications, the number of matching rules that must be authored grows with the \textit{conversational domain} (not just the task domain). In videogames, where game worlds might simulate the real world (to some level of fidelity), players who interact with conversational NPCs will expect vast conversational domains. \textit{Fa\c{c}ade}, whose NLU system is rule-based, partly wrangles this problem by constraining the domain according to a strong dramatic progression. While a player may produce any utterance at any time, he or she can recognize that even utterances that were in-domain at the beginning of gameplay, such as small talk, are effectively out-of-domain in later dramatic beats (e.g., at the story's climax). This notion even materializes architecturally, with rule subsets that are explicitly tied to specific dramatic beats \cite{mateas2004natural}. This improves the efficacy of the NLU system in those dramatic beats, but it also serves as authorial scaffolding---it is less daunting to author a series of small rulesets for salient dramatic situations than a massive ruleset for an essentially unconstrained conversational domain. Even with this scaffolding, the authoring task was immense: Mateas and Stern produced 6,800 rules over the course of hundreds of person hours (and then still relied on the additional measure of rules being promiscuous in their mapping to discourse acts) \cite{mateas2004natural}. As such, it is not surprising that, nearly fifteen years since its first reporting in the literature, very few practitioners of entertainment-based interactive media have taken on the massive authorial burden requisite to employing \textit{Fa\c{c}ade}'s demonstrated technical approach \cite{lessard2016designing}. Additionally, we note that the prospect of taking this rule-based approach would be even more daunting in interactive media lacking strong dramatic progression, e.g., open-world games. Further, the rules themselves can be difficult for naive authors---e.g., dialogue authors working on teams developing videogames---to compose. Finally, beyond authorial burden, there is the basic problem that matching rules, even fuzzy ones, are often brittle.


In this paper, we present a method for NLU that is intended to be less authorially intensive, less confounding to naive authors, and less brittle than rule-based approaches. This method utilizes \textit{context-free grammars} (CFGs) in conjunction with the \textit{long short-term memory} (LSTM) \textit{recurrent neural network} (RNN) architecture. Specifically, a (potentially naive) author specifies a CFG (using a tool we have developed called Expressionist \cite{ryan2015towardnlg}) whose terminal derivations are surface utterances and whose nonterminal symbols are annotated by the author to capture semantic and pragmatic considerations. From this CFG, we generate training data in the form of surface derivations paired with traces of the grammatical expansions that produced them. The learning task, then, is one of \textit{sequence-to-sequence} (seq2seq) translation, in which we train an RNN to map from surface derivations to grammatical traces. Crucially, because the symbols in these traces have been annotated with semantic and pragmatic information, we can infer such information from any trace that the RNN translates a surface utterance into. 

We are currently employing this method in a game that we are developing, called \textit{Talk of the Town} \cite{ryan2015towardchars}, by having a trained RNN translate arbitrary player utterances to grammatical traces, which are then used to procure semantic and pragmatic information that is fed to the game's dialogue manager. While we are not yet poised to explicitly compare our method to rule-based systems in terms of authorial burden, amenability to naive authors, or brittleness, we do demonstrate its accuracy in translating from surface derivations to grammatical traces (which point directly to semantic and pragmatic mark-up); additionally, we will attempt to qualitatively argue for the advantages of our approach, relative to rule-based systems, along those criteria. Finally, we also provide examples of actual in-game conversations, using the method we describe herein, between a player and NPCs in \textit{Talk of the Town}. Beyond the general contribution of demonstrating a new NLU approach combining CFGs with deep learning, this is to our knowledge the first usage of seq2seq learning for conversational agents who explicitly reason over semantic and pragmatic considerations.

\section{Related Work}

As noted above, our targeted application is freeform conversation between player and NPCs in a videogame. Beyond \textit{Fa\c{c}ade} \cite{mateas2004natural}, discussed in this previous section, there have been at most a handful of released titles in this area that have supported such interaction \cite{lessard2016designing}. To our knowledge, these games have all employed rule-based approaches.

A large amount of previous work in the creation of conversational agents has relied on a lengthy processing pipeline that proceeds from surface text through syntactic analysis, semantic analysis, pragmatic analysis, and dialogue management before finally producing a response \cite{ConversationalAgents}. Not all of these steps are always employed, however: the creation of a chat agent can be as straight-forward as simple string pattern matching rules \cite{ChatScript}, but can also involve a number of preprocessing steps such as part-of-speech tagging, lemmatization, sentiment analysis \cite{wong2012flexible}, and stemming \cite{chakrabarti2012semantic}. At their core, though, these approaches all rely on pattern matching (some just have more finely grained patterns), and require successful authoring of both patterns and corresponding meaning representations.

Other approaches have been considered to help alleviate the authorial burden for conversational systems.  O'Shea et al. \cite{o2008novel} used WordNet to calculate distances between words and then used a weighted sum of word distance and word ordering as an automatic method for determining similarity between a user's utterance and a sentence associated with a rule.  This still requires a designer to author rules and associated sentences, but alleviates the pattern matching authoring.

Sequence-to-sequence (seq2seq) learning is a framework first put forth by Sutskever et al. \cite{sutskever2014sequence}. These systems represent the current state of the art for many target domains such as grammatical sentence parsing \cite{vinyals2015grammar}, simulating the execution of simple programs \cite{zaremba2014learning}, or generating code to enact cards from Hearthstone and Magic: The Gathering \cite{LPN}. Vinyals and Le \cite{vinyals2015neural} used seq2seq learning to train a chatbot from a corpus of movie subtitles; this system is very impressive, and is capable of performing very well locally, but has no long term memory. With memory of the conversation only implicitly encoded in the RNN, it is highly likely that the system will not remember what it has previously said.  

It also has the problem that similar, but not identical, questions can have drastically different results, such as:
\begin{displayquote}
\textbf{Human}: what is your job ? \\
\textbf{Machine}: i'm a lawyer . \\
\textbf{Human}: what do you do ? \\
\textbf{Machine}: i'm a doctor .
\end{displayquote}
With memory of the conversation only implicitly encoded in the RNN, it is highly likely that the system will not remember what it has previously said.

\section{Motivation}

The target domain for this work is \textit{Talk of the Town} \cite{ryan2015towardchars}, an asymmetric multiplayer simulation-based game focused on navigating NPC social networks and exploring their subjective world knowledge to either hide or determine a player's identity. The NPCs are all residents of a small town in the late 1970s and the player's main mode of interaction is to converse with them naturally.  To do so, the player must type their utterances, and the game must be able to interpret these utterances to allow NPCs to understand a player's speech, query their knowledge, and respond.  The game is currently under development.


Previous iterations of the game had players navigate tedious dialogue trees to construct their conversation. This caused many players to disengage and to believe that the underlying system, a complex social network of agents with full memory models, was exceedingly simple. Wardrip-Fruin refers to this as the \textit{Tale-Spin Effect } \cite{ExpressiveProcessing} saying that ``[it] is far from uncommon in digital media, perhaps particularly in the digital arts, where fascinating processes [...] are often encased in an opaque interface." As such, a seamless interface for human conversation is needed, and this work represents a large step forward in allowing players to converse naturally with the simulated agents within the game. While elsewhere we report on our method for NLG in the game \cite{ryan2016aiide}, this paper describes our approach to NLU.


\section{Method}

Our work has three major parts, which we will discuss in turn in this section:
\begin{enumerate}
\item Using a CFG to generate surface utterances and the corresponding traces of their grammatical expansions, which may be used to collect semantic and pragmatic information.
\item Using the generated utterance--trace pairs to train a seq2seq model to map from utterance to grammatical trace.
\item Using predicted grammatical traces to collect semantics and pragmatic information, which is then passed to a dialogue manager that controls a conversational agent.
\end{enumerate}

We will now detail the data generation process, and discuss the data set used for this work.  We will then discuss the seq2seq framework, and finally outline the processing of the grammar for semantics and pragmatics. 
\subsection{Training Data}

To our knowledge there does not exist a good corpus for open-ended conversation.  Vinyals and Le used the OpenSubtitles dataset \cite{tiedemannnews} (which comprises movie subtitles) for their conversational agent, but such a dataset poses two problems, one small, the other insurmountable.  The first problem is that the set of all movie subtitles is not necessarily a good match for dialogue that would make sense in our context.  e.g., this interaction with their agent:
\begin{displayquote}
\textbf{Human}: what is the purpose of existence ? \\
\textbf{Machine}: to find out what happens when we get
to the planet earth .
\end{displayquote}
Obviously, this response has a healthy dose of science fiction, and as such would not be a good match for a small American town in the late 1970s (the setting of \textit{Talk of the Town}). While it would be possible to cull all science fiction, fantasy, and other obviously incompatible entries from the training data, this example speaks to a larger problem. Movies tend to deal with heightened situations that often shy away from the small, intimate, and somewhat mundane conversations that we seek to support in our game.

The insurmountable issue is one of annotation. We wish to be able to go from text to semantics and pragmatics, and it would require a huge effort to annotate an existing unlabeled corpus. For this, we utilize Expressionist, a tool for authoring probabilistic CFGs whose terminal derivations come furnished with mark-up \cite{ryan2015towardnlg}. Specifically, annotations (using tagsets and tags that the author defines) are attributed by an author to nonterminal symbols in the grammar. Whenever a symbol is expanded during the process of producing a terminal derivation (in our case, a surface utterance), the mark-up attributed to that symbol is accumulated; once the derivation has been terminally derived, it will thus have accumulated all the mark-up attributed to all the symbols that were expanded in the process of deriving it. Expressionist suits the challenge of annotation burden, since a core design goal in producing the tool was to make it easier to generate large amounts of annotated data.

For our purposes here, we utilize an Expressionist grammar that had already been authored for the purpose of generating NPC dialogue in \textit{Talk of the Town}. In this grammar, the mark-up attributed to nonterminal symbols corresponds to the semantic and pragmatic concerns that the game's dialogue manager operates over, which is described at length in \cite{ryan2016characters}. By using Expressionist, we are able to quickly generate large swaths of surface variation, with each individual variant being explicitly annotated for all the concerns of this dialogue manager. Specifically, our grammar took approximately twenty hours for a single author to generate, and comprises 217 nonterminal symbols and 624 production rules; in total, this grammar is capable of yielding a total 2,805,121 surface utterances.


As mentioned above, the actual training data used for training our RNN (which we describe in the next subsection), comprises pairs of surface derivations matched with traces of the grammatical expansions that produced them. Here are a few examples of such pairs, taken directly from our data:
\begin{displayquote}
\textbf{Utterance:} ``Oh, greetings, Andrew."  \\
\textbf{Trace:} \texttt{greet( greet back( use interlocutor first name ) )} \\

\textbf{Utterance:} ``I'm alright. Yourself?" \\ 
\textbf{Trace:} \texttt{answer (answer how are you (answer how are you neutrally) (ask (ask how are you (make small talk) ) ) )} \\

\textbf{Utterance:} ``Yes, the weather is wonderful." \\
\textbf{Trace:} \texttt{agree( agree about the weather( agree that the weather is good( say something positive) ( say something positive)))}
\end{displayquote}
As we have explained above, the symbols in these traces (e.g., \texttt{greet back}) have been annotated with semantic and pragmatic information, which means that \textit{Talk of the Town}'s dialogue manager can deeply reason about an utterance (e.g., ``Oh, greetings, Andrew.'') simply by collecting the mark-up attributed to all the symbols in its trace.

To produce the training data for our learning task, we specifically carried out the following steps:
\begin{enumerate}
\item Produce all 2,805,121 terminal derivations generable by our authored CFG.
\item Postprocess the full set of terminal derivations to produce a subset that is \textit{balanced} with regard to the symbolic content of the derivations' corresponding traces.
	\subitem Specifically, we randomly selected 5000 terminal derivations for each unique (unordered) set of nonterminal symbols that appeared together in at least one trace. For instance, in the first example utterance--trace pair above, the unique set of symbols appearing in the trace is \texttt{\{greet, greet back, use interlocutor first name\}}. This was done so that variations for capsules of semantic and pragmatic concerns (e.g., responding positively about the weather) would not be overrepresented in the training data. Because there were 69 such unique symbol sets that appeared across all traces, we ended up sampling a total of 345,000 utterance--trace pairs.
\item Augment these 345,000 utterance--trace pairs with new pairs whose utterances are corruptions of the original utterances (corrupted by randomly removing a third of the utterance tokens), as well as pairs whose utterance punctuation was stripped; this yielded a total set comprising 1,035,000 pairs.
	\subitem The addition of a denoising component is intended to make the system more robust to both \textit{out-of-vocab} (OOV) problems and, in conjunction with the removal of punctuation, should make our trained model more robust to terse player input (e.g., ``you from here", as opposed to the more verbose ``Are you from around here?").
\end{enumerate}


\subsection{Seq2Seq Learning Procedure}

\begin{figure*}[t]
\centering
    \includegraphics[width=0.95\textwidth]{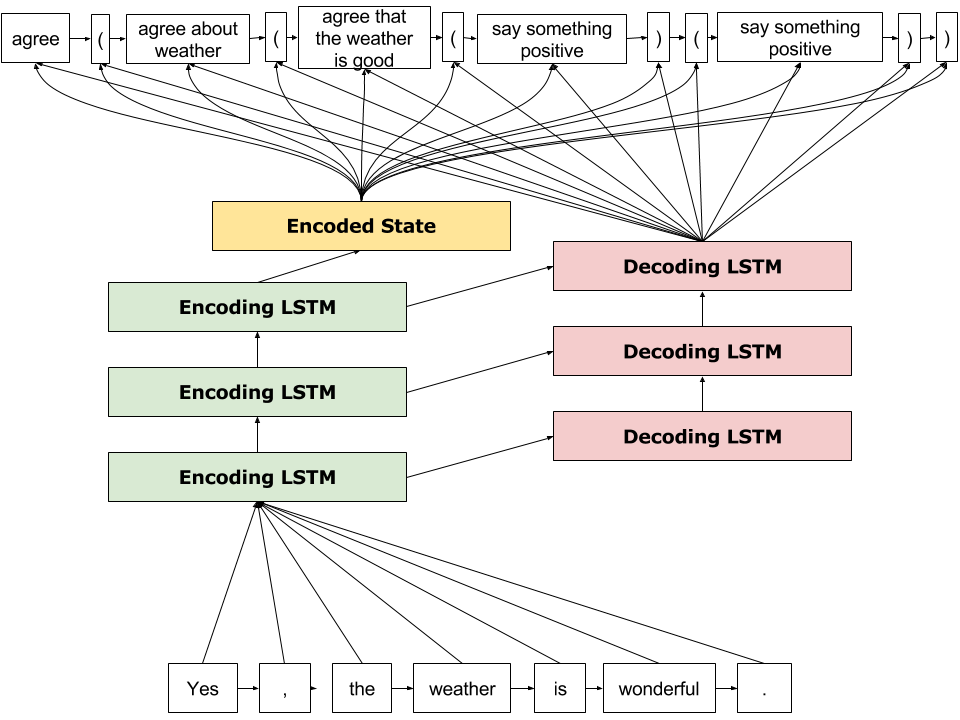}
    \caption{Illustration of our seq2seq framework.  The input sequence is seen at the bottom and is encoded via a stack of three LSTM layers to an encoded state that is then decoded via a stack of three decoding LSTM layers.  The final decoding process uses an attentional mechanism that allows it to look back at the encoded state and focus on specific aspects at different times, producing the output sequence at top.}
    \label{fig:RNN}
\end{figure*}

\begin{figure*}[ht]
\centering
    \includegraphics[width=0.50\textwidth]{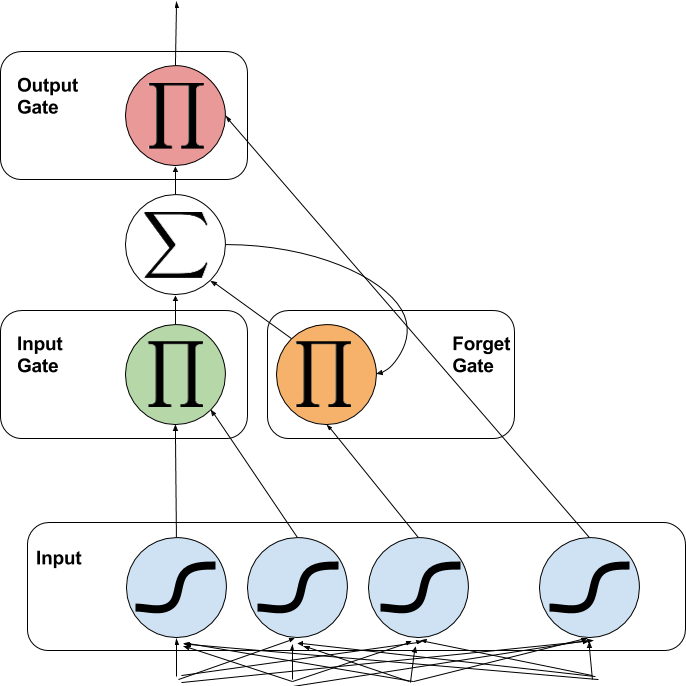}
    \caption{Illustration of an LSTM Cell. The input comes in from the bottom, from a densely connected layer (i.e., every input token has an edge to each input node) and is passed through a $tanh$ function. Each of these nodes then passes its $0-1$ output to one of three separate nodes.  Two go to the input gate (one acting as a gate, the other acting as the input).  One is passed to the forget gate that determines whether the saved value is retained or dropped. The memory is additively retained.  Finally, the memory value goes to the output gate where the final input value determines whether the value is output or not.  }
    \label{fig:LSTM}
\end{figure*}

In the section, we will describe our seq2seq learning procedure, with a particular focus on explaining how the technique generally works. For this task, we used the Tensorflow framework \cite{Tensorflow}. An illustration of our network can be seen in Figure \ref{fig:RNN}; as the figure shows, the input sequences are composed of the individual tokens (words and punctation marks) appearing our surface utterances, and the output sequences are composed of the discrete references to individual nonterminal symbols that appeared in the grammatical traces (i.e., each symbol correspond to a single token). The topology used in this work consisted of three encoding layers and three decoding layers, with each layer consisting of 384 LSTM \textit{cells}. LSTMs represent the current state of the art for sequence processing and are a modification of the standard RNN approach, first put forth by Hochreiter and Schmidhuber \cite{LSTM}.  A standard RNN modifies the feed-forward neural network by adding an additional feedback edge that modifies itself after each input, which allows the network to learn sequences.  However, a large problem with RNNs is the \textit{vanishing/exploding gradient problem}.  The multiplicative update process for the recurrent edge means that gradients change exponentially with time, which leads to either vanishing (sub-unit weights) or exploding (larger-than-unit weights) and limits the considered time-scale to approximately three to five inputs.  LSTMs circumvent this process by placing guards on the input, output, and retention for the recurrent edges.  An LSTM cell can be seen in Figure \ref{fig:LSTM}.  A crucial aspect is that the recurrent edge is updated additively, instead of multiplicatively, which eliminates the vanishing gradient problem and allows the LSTM to be trained on much longer sequences (e.g., our work utilizes sequences of up to 80 tokens).

The encoding stack processes the input to an intermediate hidden vector. This hidden vector is then passed to the decoding LSTM as an initial state along with a special \textit{START} token. As each token is decoded, the state is updated via the LSTM recurrent cells, but the decoder is allowed to look back at the original encoded state.  It does so with a soft-attention mechanism that computes a softmax distribution over the hidden state that acts as a mask, albeit not a hard mask (as the mask is constrained to sum to 1 and values will be between 0 and 1). Decoding is handled in a greedy fashion with the locally maximum likelihood token being selected at each point in the output sequence.  

\subsection{Online NLU During Gameplay}

After training the RNN, we incorporated it into the software framework that underpins \textit{Talk of the Town}. As described in \cite{ryan2016characters}, conversation in our game is turn-based, with turns being allocated by the dialogue manager. When a turn has been given to a player character, the player is asked to furnish her character's next utterance. Once the player has submitted this, the dialogue manager passes the utterance to the RNN, which tokenizes it and performs seq2seq translation on it to produce a grammatical trace composed of symbols in our Expressionist grammar. From here, the dialogue manager collects all the mark-up associated with all the symbols appearing in the trace, and uses this to update the conversation state. From here, the next turn will be allocated---potentially to the NPC, whose line will be generated according to the conversation state---and the conversation will proceed in this way until completion. For OOV input (i.e., player-produced tokens that did not appear in the training data), we employ a short pipeline that utilizes a spellchecker and, if necessary, the Google News \textit{word2vec} model \cite{GoogleNewsW2V} to convert to the semantically closest in-vocab token. 


We amortized the word2vec computation by precomputing the semantically closest word in the training vocabulary for each of the 3M words in the word2vec model. When an OOV token is encountered, the following procedure is enacted:

\begin{algorithm}
\begin{algorithmic}[1]
\Procedure{}{}
\For{top spell-check candidates}
\If{candidate in vocab}
\State return candidate
\EndIf
\If{candidate in word2vec}
\State return closest in vocab
\EndIf
\EndFor
\State return \textit{OOV}
\EndProcedure
\end{algorithmic}
\end{algorithm}

A character-level input model was considered, but ultimately decided against.  While it would eliminate the concept of OOV tokens, gaps in the vocabulary would still exist.  Consider a player asking, ``Have you seen a guy with auburn hair?". Unless `auburn' exists in the vocabulary, a character-level model will have very little information to go off of. It would probably know that auburn was a color via context, but what color it chose in its stead would be impossible to determine a priori. Instead, with the word2vec mapping, `auburn' would be mapped to `brown', which the system would be able to reason about properly.   
\begin{table}[t]
\begin{center}
\centering
\begin{tabular}{| c | c | c | c |}
\hline
\textbf{Split} & \textbf{Cross Validation Perplexity} & \textbf{Test Perplexity} \\
\hline 
1 & 1.046 & 1.047 \\
2 & 1.048 & 1.045 \\
3 & 1.043 & 1.044 \\
4 & 1.053 & 1.053 \\
5 & 1.046 & 1.047 \\
6 & 1.044 & 1.043 \\
7 & 1.045 & 1.044 \\
8 & 1.042 & 1.044 \\
9 & 1.048 & 1.048 \\
10 & 1.043 & 1.043 \\
\hline
\end{tabular}
\end{center}
\caption{Perplexity values for our RNN's performance on both a cross-validation task ($n=846,820$) and using a held-out test set ($n=94,090$) for each fold ($n=94,090$).}
\label{table:results}
\end{table} 

\section{Evaluation}

In this section, we present the promising results of an offline evaluation experiment that demonstrates the accuracy of our system in mapping from surface utterances to grammatical traces. To conduct this experiment, we randomized our set of training data ($n=1,035,000$), split it into eleven pieces ($n=94,090$), held one out as a test set, and for each of the other ten pieces, performed 10-fold cross validation on the remainder of the non-held-out set ($n=846,820$) before finally using the held-out piece as a test set. The results, shown in Table \ref{table:results} using \textit{perplexity} values, indicate that this approach is robust to variations and gaps in the data, with no fold performing drastically better or worse than any other. Perplexity, a measure of how well a probability distribution is able to predict a given sample, is defined as
$b^{- \frac{1}{N} \sum_{i=1}^N \log_b q(x_i)}$, where $b$ is typically 2, $x_1 \dots x_N$ are the samples, and $q$ is the probability distribution.  A perplexity of $k$ effectively means that there were $k$ many equally valid choices according to the probability distribution.  For instance, a fair coin would have perplexity of 2 and a fair six-sided dice would have perplexity of 6.  Perplexity values near 1 showcase the ability of the system to translate from surface utterances to grammatical traces (and thereby semantic and pragmatic information, as explained above) nearly perfectly.  Given that this holds on both the cross validation and held out data, we are satisfied that this high accuracy is not the result of overfitting, especially considering the fact that we corrupted one third of the surface utterances in our training data. In tandem with our module for resolving OOV tokens, we feel that the system may prove robust to player dialogue constructions that differ from variants that appear in the training data; of course, the substantiation of this in an online evaluation study (with real player inputs) remains as future work. 

Given that these nearly perfect perplexities come from corrupted utterances a third of the time, it appears that the system may be robust to sentence construction choices and misunderstandings that arise from  OOV word repair process detailed in section 4.1.

\subsection{Example Conversations}

As informal qualitative evidence of system performance, we will present here two conversations produced by having an author of this paper engage in dialogue with an NPC during actual gameplay. We should note first that named-entity recognition here is currently hard coded as a symbol referencing the speaker's name (e.g., ``I'm $<$SPEAKER$>$, by the way" instead of ``I'm Ben, by the way").

The current topics of conversation that can be represented are:
\begin{itemize}
\item Introductions
\item Farewells
\item The weather
\item Where someone is from
\item Where someone works
\item Pleasantries
\end{itemize}

~\newline
\textbf{CONVERSATION 1: Initial meeting of Joe Imig-Bader (player) and Susan Despres-Trumbauer (NPC) at Schmitz Diner on the night of August 16, 1979.}

\begin{displayquote}
\textbf{[P] Joe Imig-Bader:} Hello.

\textbf{[NPC] Susan Despres-Trumbauer:} Oh, hi!

\textbf{[NPC] Susan Despres-Trumbauer:} This weather is amazing.

\textbf{[P] Joe Imig-Bader:} It's spectacular!

\textbf{[NPC] Susan Despres-Trumbauer:} Really, though: It's amazing.

\textbf{[P] Joe Imig-Bader:} It really is nice.

\textbf{[P] Joe Imig-Bader:} How are you doing?

\textbf{[NPC] Susan Despres-Trumbauer:} Fine. And yourself?

\textbf{[P] Joe Imig-Bader:} Fine, fine.

\textbf{[NPC] Susan Despres-Trumbauer:} I'm Susan, by the way.

\textbf{[P] Joe Imig-Bader:} I'm Joe.

\textbf{[NPC] Susan Despres-Trumbauer:} Nice to meet you.

\textbf{[P] Joe Imig-Bader:} Nice to meet you.

\textbf{[NPC] Susan Despres-Trumbauer:} What do you do?

\textbf{[P] Joe Imig-Bader:} I don't work.

\textbf{[P] Joe Imig-Bader:} What do you do?

\textbf{[NPC] Susan Despres-Trumbauer:} I work at 8th Street Delicatessen.

\textbf{[P] Joe Imig-Bader:} Sorry, this is rude, but I have to go.

\textbf{[NPC] Susan Despres-Trumbauer:} Goodbye.

\textbf{[P] Joe Imig-Bader:} Bye.

\end{displayquote}

\noindent \textbf{CONVERSATION 2: Initial meeting of Barbara Perrodin (NPC) and Ben Lashley (player) at 9th Street Delicatessen on the night of August 13, 1979.}

\begin{displayquote}
\textbf{[NPC] Barbara Perrodin:} Hey.

\textbf{[P] Ben Lashley:} Oh, hi.

\textbf{[P] Ben Lashley:} How's it going?

\textbf{[NPC] Barbara Perrodin:} Not bad.

\textbf{[NPC] Barbara Perrodin:} How's it going with you?

\textbf{[P] Ben Lashley:} Pretty good.

\textbf{[P] Ben Lashley:} Do you work around here?

\textbf{[NPC] Barbara Perrodin:} I work at Law Offices of Foeller \& Ansel-Gumbel.

\textbf{[P] Ben Lashley:} You don't look familiar, though. Are you from this town?

\textbf{[NPC] Barbara Perrodin:} Nope, I wasn't born here.

\textbf{[P] Ben Lashley:} I'm Ben, by the way.

\textbf{[NPC] Barbara Perrodin:} I'm Barbara.

\textbf{[P] Ben Lashley:} It's nice to meet you.

\textbf{[NPC] Barbara Perrodin:} Nice to meet you.

\textbf{[P] Ben Lashley:} Well, I better run now. Nice chatting with you.

\textbf{[NPC] Barbara Perrodin:} Goodbye.

\textbf{[P] Ben Lashley:} Bye.

\textbf{[NPC] Barbara Perrodin:} Bye.
\end{displayquote}

\textbf{CONVERSATION 3: Louis Ranger (player) and Sandra Dreher (NPC) at Chez Maggie on the night of August 16, 1979.}

\begin{displayquote}
\textbf{[P] Louis Ranger:} Hi there!

\textbf{[NPC] Sandra Dreher:} Hi there, Jack.

\textbf{[P] Louis Ranger:} That's not my name.

\textbf{[NPC] Sandra Dreher:} Oops, not sure why I thought that.

\textbf{[P] Louis Ranger:} Hey, no worries!

\textbf{[NPC] Sandra Dreher:} How's it going?

\textbf{[P] Louis Ranger:} Not too bad. You?

\textbf{[NPC] Sandra Dreher:} Good.

\textbf{[NPC] Sandra Dreher:} This weather is good.

\textbf{[P] Louis Ranger:} It's great!

\textbf{[P] Louis Ranger:} I'm Louis, by the way.

\textbf{[NPC] Sandra Dreher:} Sandra.

\textbf{[P] Louis Ranger:} Nice to meet you!

\textbf{[NPC] Sandra Dreher:} Nice to meet you.

\textbf{[P] Louis Ranger:} I should run. Maybe I'll see you around here again.

\textbf{[NPC] Sandra Dreher:} Bye.

\textbf{[P] Louis Ranger:} See ya around.
\end{displayquote}

\section{Discussion and Future Work}

In this paper, we have presented a novel approach to NLU that utilizes context-free grammars in conjunction with sequence-to-sequence deep learning. Specifically, we take a CFG that is capable of generating dialogue for our target application \cite{ryan2016aiide}, a game called \textit{Talk of the Town} \cite{ryan2015towardchars}, and train a long short-term memory recurrent neural network to map from the surface utterances that it produces to traces of the grammatical expansions that derived them. Critically, this CFG was authored using a tool we have developed called Expressionist \cite{ryan2015towardnlg}, which supports arbitrary annotation of the nonterminal symbols in the grammar. Because we annotated the symbols in this grammar for the semantic and pragmatic considerations that our game's dialogue manager \cite{ryan2016characters} operates over, we can use the grammatical trace associated with any surface utterance to infer such information. During gameplay, we translate player utterances into grammatical traces (using our RNN), collect the mark-up attributed to the symbols included in that trace, and pass this information to the dialogue manager, which updates the conversation state accordingly.

While above we demonstrated the accuracy of this system in mapping surface utterances to grammatical traces (and thereby semantic and pragmatic information characterizing the utterances), we would like to informally discuss the advantages of our method relative to rule-based approaches to NLU. First, we believe that our approach incurs less authorial burden, simply by virtue of the combinatorial explosion that characterizes generative grammars. This is demonstrated in the large number of terminal derivations that our grammar can generate, and moreover in recent additions to the grammar that have yielded on the order of \textit{quadrillions} of terminal derivations from only a few hours of authoring. Additionally, our approach is appealing in that the very same CFG may be used for both NLG \cite{ryan2016aiide} and NLU. While earlier work has explored the use of the same CFG to both parse and generate \cite{kay1996chart}, our approach here is different---rather than \textit{parse} with the CFG we have authored, we instead use its terminal derivations as actual training data, since they are annotated. This allows us to build an RNN decoder that is inherently more robust than a CFG parser, since it learns intermediate representations that are not directly encoded in the grammar's production rules. Further, we contend that our approach is more amenable to naive authors who might like to feature NLU in their applications. Rather than authoring procedural rules, by our approach an author uses the Expressionist graphical user interface, which is designed for naive authors and supports live feedback showing terminal derivations and their corresponding annotations \cite{ryan2015towardnlg}. In this sense, they author high-level patterns for how utterances should be tagged for semantic and pragmatic information, and a neural network then infers mapping rules of arbitrary granularity. While training a neural network is certainly not practical for naive authors, we plan to support black-box RNN training as a service associated with Expressionist. This way, a development team might feature naive authors who specify CFGs and other developers who specify how a dialogue manager should interface with an RNN that has already been trained. Finally, we posit that intuitively our model should be less brittle than rule-based systems. While rules in these systems work by matching discrete authored patterns (which of course may be fuzzy) against user utterances, a neural network does something similar, but with patterns at arbitrary granularities and  hierarchies (patterns of patterns) that are learned dynamically. Of course, one tradeoff here is that human-authored rulesets are much more interpretable than RNNs.



A more robust handling of names (and other specifics mentioned in the data) is something we intend to cover in future work.  For the type of small-talk focused conversation currently modeled by the system, there are not a lot of specifics that need to be filled in, but future iterations will need to be able to answer targeted questions (e.g., ``Do you know Ben Lashley?'').  Ling et al. \cite{LPN} introduced an extension of the seq2seq framework called \textit{Latent Predictor Networks}. Instead of just being able to generate a single token from a vocabulary at decoding time, they are capable of choosing a number of different actions based on a latent predictor that chooses whether to perform the standard LSTM token generation, copy from a structured input field, or, most importantly, copy a token from the input field.  Copying a token from the input field using a Pointer Network \cite{vinyals2015pointer} is exactly what is needed to accurately translate the specifics of questions players may ask in our game (e.g., specific character attributes or names).

%
%
%

While this system is capable of producing a large amount of surface text generation, there is remaining authoring work to be done to expand the scope of generable dialogue. Finally, while our own initial interactions with the system have been successful, we are currently planning a study with actual players so that we may better understand both the successes and limitations of the system.

 Another focus for future work is moving away from greedy decoding scheme to a more globally aware decoding scheme, such as a beam search, to search a wider space of output sequences.  Similarly, to improve the translation process we would like to incorporate the conversational history into the input. Currently, the system only considers a user's utterance alone, with no notion of the surrounding context.  This is typically handled correctly by the simulated agent's dialogue management, but in future work we would like to include previous turns of the conversation in the input stream to better handle ambiguities that can arise in the translation process.

\bibliographystyle{splncs03}
\bibliography{bibliography} 
\end{document}